\pdfoutput=1

\documentclass[11pt]{article}

\usepackage[table,xcdraw]{xcolor}
\usepackage[]{acl}

\usepackage{times}
\usepackage{latexsym}

\usepackage[T1]{fontenc}

\usepackage[utf8]{inputenc}

\usepackage{microtype}
\usepackage{subfigure}
\usepackage{graphicx}
\usepackage{cleveref}
\usepackage{makecell}
\usepackage{multirow}
\usepackage{booktabs}
\usepackage{arydshln}
\usepackage{rotating}

\newcommand*\samethanks[1][\value{footnote}]{\footnotemark[#1]}

\title{On the Importance of Data Size in Probing Fine-tuned Models}

\author{Houman Mehrafarin\thanks{$~~$The authors contributed equally to this work.}$~~^{\diamondsuit}$\hspace{1em} Sara Rajaee\samethanks$~~^{\diamondsuit}$ \hspace{1em} Mohammad Taher Pilehvar$~^\spadesuit$\\
  $~^{\diamondsuit}$Iran University of Science and Technology, Tehran, Iran \\
  $~^\spadesuit$Tehran Institute for Advanced Studies, Khatam University, Iran \\
  \texttt{\{h{\_}mehrafarin, sara{\_}rajaee\}@comp.iust.ac.ir}\\
  \texttt{mp792@cam.ac.uk}}

\begin{document}
\maketitle
\begin{abstract}

Several studies have investigated the reasons behind the effectiveness of fine-tuning, usually through the lens of probing. 
However, these studies often neglect the role of the size of the dataset on which the model is fine-tuned.
In this paper, we highlight the importance of this factor and its undeniable role in probing performance.
We show that the extent of encoded linguistic knowledge depends on the number of fine-tuning samples.
The analysis also reveals that larger training data mainly affects higher layers, and that the extent of this change is a factor of the number of iterations updating the model during fine-tuning rather than the diversity of the training samples.
Finally, we show through a set of experiments that fine-tuning data size affects the recoverability of the changes made to the model's linguistic knowledge.\footnote{We have released our code and models' checkpoints at: \url{https://github.com/hmehrafarin/data-size-analysis}}

\end{abstract}

\section{Introduction}

The outstanding performance of pre-trained language models (LMs) on many NLP benchmarks has provoked curiosity about the reasons behind their effectiveness. 
To this end, several probes have been proposed to explore their capacity \citep{47786, hewitt-manning-2019-structural, wu-etal-2020-perturbed}. 
The investigations have clearly highlighted the ability of LMs in capturing various types of linguistic knowledge \citep{liu-etal-2019-linguistic,clark-etal-2019-bert,michael-etal-2020-asking,klafka-ettinger-2020-spying,tenney-etal-2019-bert}. 

However, to take full advantage of the encoded knowledge of pre-trained models in specific target tasks, it is usually required to perform a further fine-tuning \citep{devlin-etal-2019-bert}.  The broad application of fine-tuning has garnered the attention of many researchers to explore its peculiarities.  Trying to understand the fine-tuning procedure, recent analyses have shown that most of the pre-trained linguistic knowledge is preserved after fine-tuning \citep{47786}.  Furthermore, by encoding the essential linguistic knowledge in the lower layers, this procedure makes the higher layers task-specific \citep{durrani-etal-2021-transfer}.  However, \citet{mosbach-etal-2020-interplay-fine} argued that the changes in the probing performance can not be attributed entirely to the modifications a model undergoes with respect to its linguistic knowledge after fine-tuning.

While the previous studies focused on the role of the target task as a factor that affects the probing performance of fine-tuned models, we present another important factor in interpreting probing results for such models.  Our investigations reveal that the conclusions drawn by previous probing studies that investigate the impact of fine-tuning on acquiring or forgetting knowledge might not be entirely reliable unless the size of the fine-tuning dataset is also taken into account.  Through several experiments, we show that the encoded linguistic knowledge can highly depend on the size of target tasks' datasets.  Specifically, the larger the task data, the more the probing performance deviates from the pre-trained model, irrespective of the fine-tuning tasks. 

To address the overlooked role of data size, we run several experiments by limiting training samples and probing the fine-tuned models.  Our results indicate that models fine-tuned on large training datasets witness more change in their encoded linguistic knowledge compared to pre-trained BERT.  However, by reducing fine-tuning training data size (e.g., from 393k in MNLI to 7k), the gap between probing scores becomes smaller.  Moreover, we expand our analysis and evaluate the extent to which large training datasets affect the captured knowledge across layers.  The layer-wise results show that the effect of data size is more notable on higher layers, particularly for models trained on larger datasets.  We take our analysis a step further and show that the difference in probing performance among different data sizes is due to the total number of optimization steps rather than the diversity of training samples.  Finally, through a set of experiments, we show that the changes made to the probing performance by a fine-tuning task can be recovered if the model is re-fine-tuned on a task with comparable data size.

The findings of this paper can be summarized as follows:
\begin{itemize}
    \item Data size is a factor that highly impacts a fine-tuned model's probing performance.
    
    \item The size of the dataset mainly affects the probing performance of the higher layers.
    
    \item The number of training steps is what makes larger datasets have higher impacts on the model's linguistic knowledge (rather than the diversity in training samples).

    \item {Fine-tuning data size affects the extent to which the modifications made to a model's linguistic knowledge are recoverable.}

\end{itemize}

\section{Related Work}

Recently, many studies have shown that pre-trained language models, such as BERT \citep{devlin-etal-2019-bert}, encode certain linguistic knowledge in their internal representations \citep{47786}. For instance, \citet{hewitt-manning-2019-structural} found that syntactic dependencies can be obtained from BERT's token embeddings, suggesting that BERT encodes syntactic knowledge in its representations. Nevertheless, not all layers behave similarly in capturing linguistic features: lower layers tend to encode surface-level knowledge, middle layers seem to be responsible for syntactic information, and higher layers capture semantic knowledge in their representations \citep{jawahar-etal-2019-bert}.

While models such as BERT capture considerable amounts of linguistic features, one still requires to fine-tune them to take full advantage of their potential in specific downstream tasks \citep{wang-etal-2018-glue}. Fine-tuning affects BERT in various ways; for instance, \citet{hao-etal-2020-investigating} found that fine-tuning mainly affects the attention mode of the higher layers and alters the feature extraction mode of the middle and last layers. In addition, fine-tuning BERT on a negation scope task improves the model's attention sensitivity to negation \citep{zhao-bethard-2020-berts}.

Apart from the changes made to BERT's attention, recent work has studied how fine-tuning affects BERT's representations and, as a result, its linguistic knowledge. \citet{merchant-etal-2020-happens} found that fine-tuning primarily affects the representations in higher layers, and depending on the downstream task, the changes made to lower layers could be either deep or shallow. Moreover, on only a small number of downstream tasks, fine-tuning seems to have a positive impact on the probing accuracy \citep{mosbach-etal-2020-interplay-fine}. Given the fact that fine-tuning mostly affects higher layers, \citet{durrani-etal-2021-transfer} showed that after fine-tuning, most of the model's linguistic knowledge is transferred to lower layers to reserve the capacity in the higher layers for task-specific knowledge.

Studies so far have relied on probing accuracy to explain how fine-tuning affects a model's linguistic knowledge \citep{mosbach-etal-2020-interplay-fine, durrani-etal-2021-transfer, merchant-etal-2020-happens}. However, given the fact that fine-tuning tasks do not share the same number of samples, concluding to what extent target tasks contribute to the model's linguistic knowledge is not fully reliable. To the best of our knowledge, none of the previous studies have considered the role of data size in fine-tuned models' linguistic knowledge. In this work, we show that the size of the dataset plays a crucial role in the amount of knowledge captured during fine-tuning. By designing different experiments, we analyze the effect of the size of the dataset in-depth.

\section{Experimental Setup}

We have carried out over 600 experiments to study the linguistic features captured during fine-tuning. This allows us to examine how much different factors impact performance on various probing tasks.
Moreover, varying the sample size lets us understand its importance in analyzing fine-tuned models.
In this section, we provide more details on setups, downstream tasks, and probing tasks.
\subsection{Fine-tuning}
\label{sec:task-size}
\label{sec:fine-tuning-setup}
For our analyses, we concentrate on the BERT-base model, which is arguably the most popular pre-trained model. 
We fine-tuned the 12-layer BERT on a set of tasks from the GLUE Benchmark \citep{wang-etal-2018-glue} for five epochs and saved the best checkpoint based on performance on the validation set.
We used the [CLS] token for classification and set the learning rate as $5e^{-5}$. 
We have chosen the following target tasks:

\paragraph{CoLA.} The Corpus of Linguistic Acceptability is a binary classification task in which \textbf{8.5k} training samples are labeled based on their grammatical correctness \citep{10.1162/tacl_a_00290}.

\paragraph{MRPC.} The Microsoft Research Paraphrase Corpus includes \textbf{3.6k} training sentence pairs in which the semantic equivalence of two sentences is determined \citep{dolan-brockett-2005-automatically}.

\paragraph{SST-2.} The Stanford Sentiment Treebank is a sentiment classification task containing \textbf{67k} training sentences \citep{socher-etal-2013-recursive}.

\paragraph{QQP.} With \textbf{364k} question pairs, the goal of the Quora Question Pairs dataset is to determine whether two questions in a pair are semantically similar.

\paragraph{MNLI.} The Multi-Genre Natural Language Inference is a Natural Language Inference (NLI) task with about \textbf{393k} records in its training set \citep{williams-etal-2018-broad}.

\begin{figure*}[!ht]
\includegraphics[width=13cm]{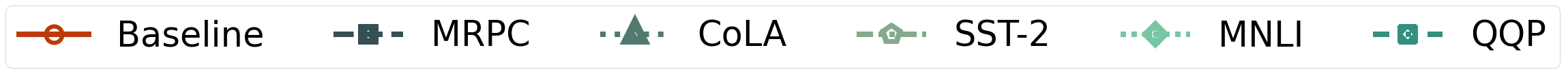}
    \centering
    \includegraphics[width=14cm,height=11cm]{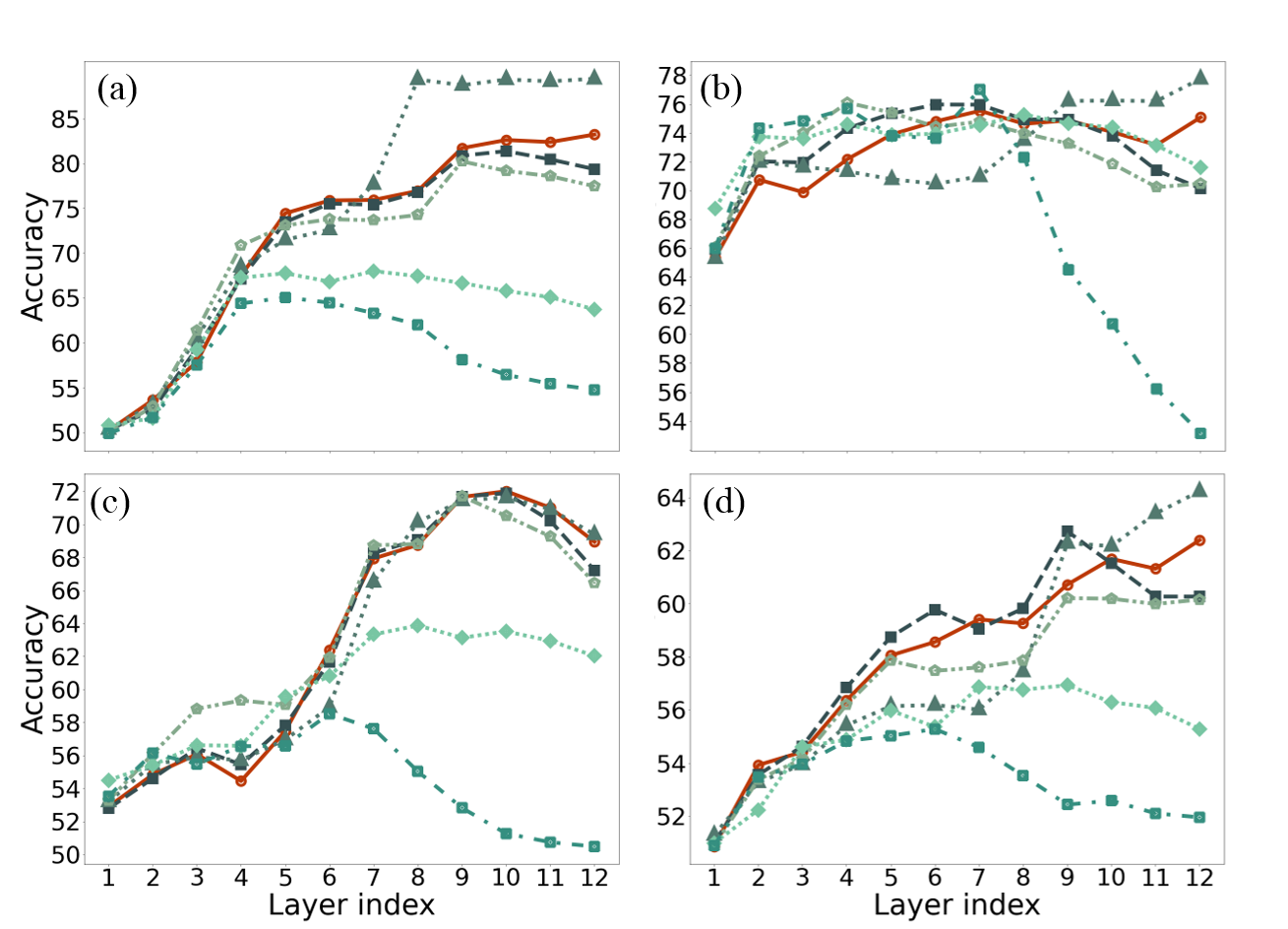} \label{fig:ling-bish}

    \caption{Probing accuracy on all the layers of fine-tuned models on (a) Bigram Shift (b) Object Number (c) Coordination Inversion (d) Semantic Odd Man Out. As shown, there is a large accuracy gap between models fine-tuned on larger data sizes (e.g., MNLI and QQP) and the baseline.}
    \label{fig:fig-probe-ling}
    
\end{figure*}

\begin{table}[t!]
\centering
\setlength{\tabcolsep}{11pt}
\scalebox{0.85}{
\begin{tabular}{l c c c c }
\toprule
  & \textbf{Full} & \textbf{7k} & \textbf{2.5k} & \textbf{1k}     \\ 
\midrule
\multirow{1}{*}{\bf CoLA}  &  57.55	 &   56.87  &	46.68	& 42.72 \\

\multirow{1}{*}{\bf SST-2} & 92.78	 &   91.28	&   89.79	& 86.81 \\

\multirow{1}{*}{\bf MNLI}  & 83.19	 &   73.73	&   68.63	& 60.16 \\
            
\multirow{1}{*}{\bf QQP}   & 90.63	 &   82.37	&   79.93	& 76.93 \\
            
\multirow{1}{*}{\bf MRPC} & 86.43	 &    -     &	81.78	& 77.82 \\

\bottomrule

\end{tabular}
}
\caption{\label{statistics-table}The performance of fine-tuned BERT on five tasks from GLUE (dev set) after fine-tuning on training data of varying size. The numbers are reported based on accuracy for SST, MNLI, QQP, MRPC, and Matthew's correlation for CoLA.}
\end{table}

\subsection{Fine-tuning performance}

The performance of the fine-tuned models on these tasks is presented in Table~\ref{statistics-table}. We report the results on different training data sizes\footnote{Since MRPC only has 3.6k training samples, we do not report any 7k results for this dataset.} to highlight the extent to which reducing training data affects a model's performance on the corresponding tasks.
It is worth mentioning that even though the performance of target tasks decreases by reducing their training data, it is still far better than the pre-trained version. Therefore, the models have learned the corresponding target tasks to some extent.

\subsection{Probing tasks}

We probe the pre-trained and fine-tuned BERT models by training a linear classifier on top while the weights of the encoders are frozen. 
Keeping the probing classifier simple allows us to scrutinize the linguistic knowledge by eliminating the possibility of the classifier learning such knowledge.
All probes are trained with a batch size of 32, a learning rate of $3e^{-4}$, a linear scheduler for adjusting the learning rate with $10\%$ warm-up steps, and for ten epochs. We also used Adam as the optimizer.
Due to limited computational resources, we were not able to run all the experiments multiple times with different random seeds.
However, to ensure the reliability of our results, we repeated several randomly chosen experiments three times (with different random seeds). 
The probing accuracy remained stable, ranging within $\pm 1.0$.
Finally, we report the evaluation scores on test sets for the models with the highest validation accuracy on the validation set.

We opted for four syntactic and semantic probing tasks from the SentEval benchmark \citep{conneau-kiela-2018-senteval} to study the linguistic knowledge encoded in the models\footnote{We also repeat our experiments on the structural probe of \citet{hewitt-manning-2019-structural}. This probe investigates how well syntactic dependency trees are encoded within a model's representations. We report the UUAS score for the distance between word pairs in the parse tree. The results are reported in Appendix~\ref{sec:struct-probe}. We choose this probe because it is different from SentEval's probes in terms of training objective to show our statement still stands.}. The binary classification tasks are as follows:

\paragraph{Bigram Shift} is a task that aims to test the model's ability to predict whether two successive random tokens in the same sentence have been inverted. 

\paragraph{Object Number} focuses on the model's ability to determine the singularity or plurality of the main clause's direct object.

\paragraph{Coordination Inversion} examines the model's ability to distinguish between original sentences and sentences where the order of two coordinated clausal conjoints have been inverted.

\paragraph{Semantic Odd Man Out} is a task that tests the model's ability to predict if a sentence is original or whether a random word has been replaced with another word from the same part of speech.

\begin{figure*}[!ht]
\includegraphics[width=13cm]{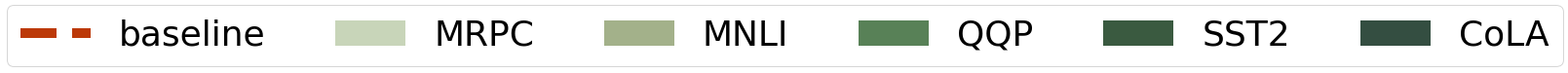}
    \centering
    \subfigure[Bigram Shift]{\includegraphics[width=7cm,height=5cm]{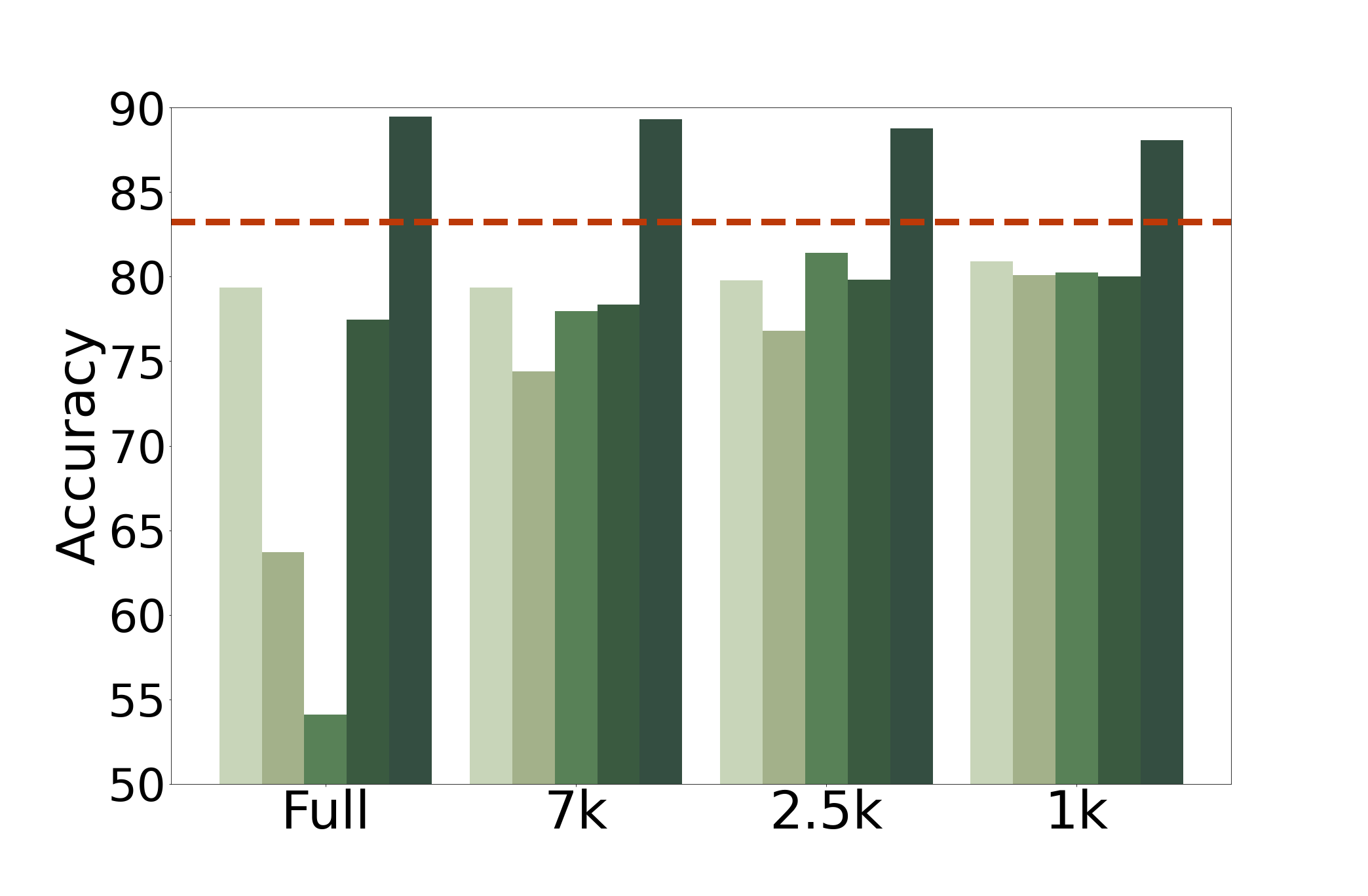} \label{fig:datasize-bish}} 
    \subfigure[Object Number]{\includegraphics[width=7cm,height=5cm]{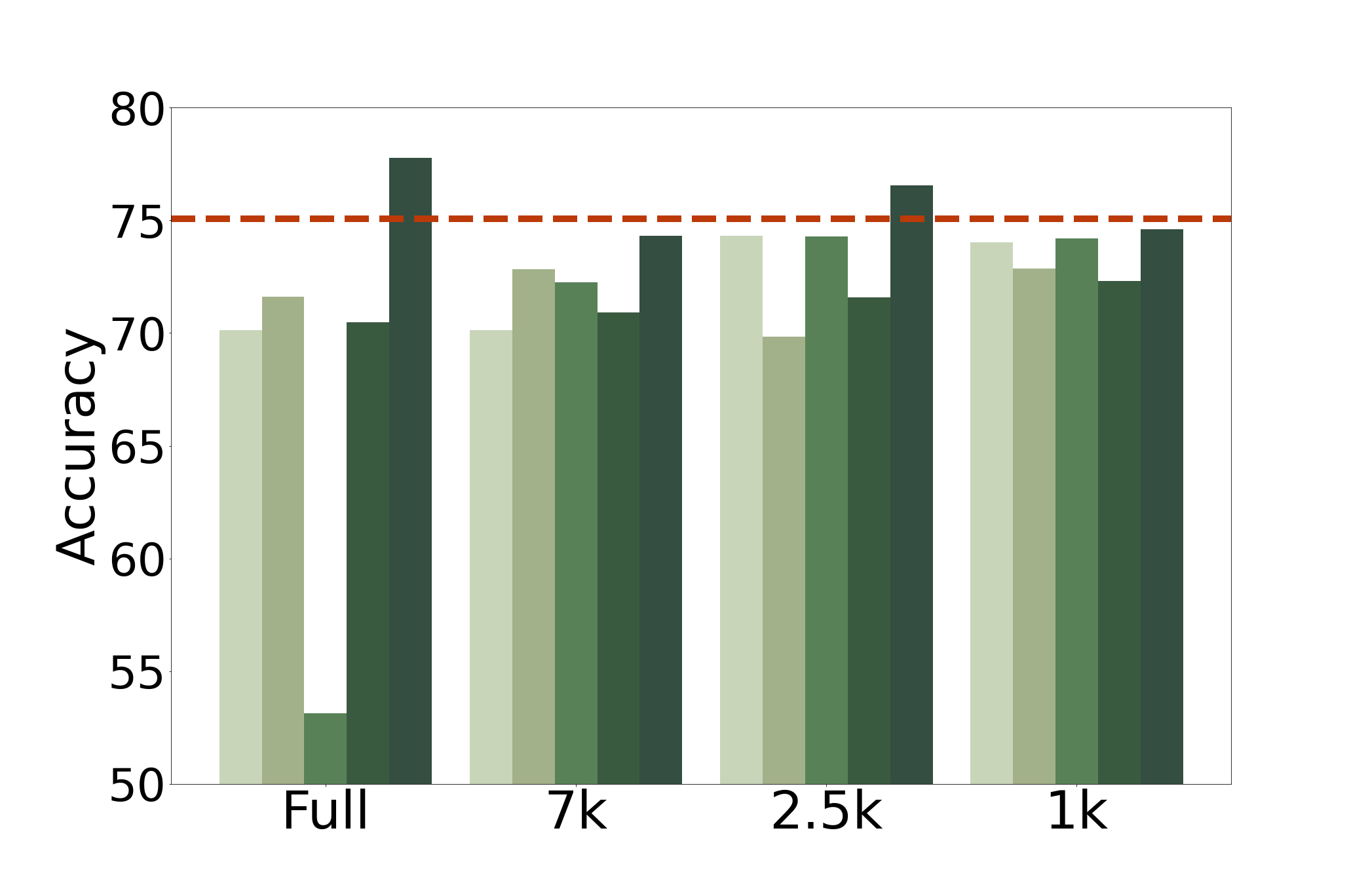}\label{fig:datasize-objnum}}
    \subfigure[Coordination Inversion]{\includegraphics[width=7cm,height=5cm]{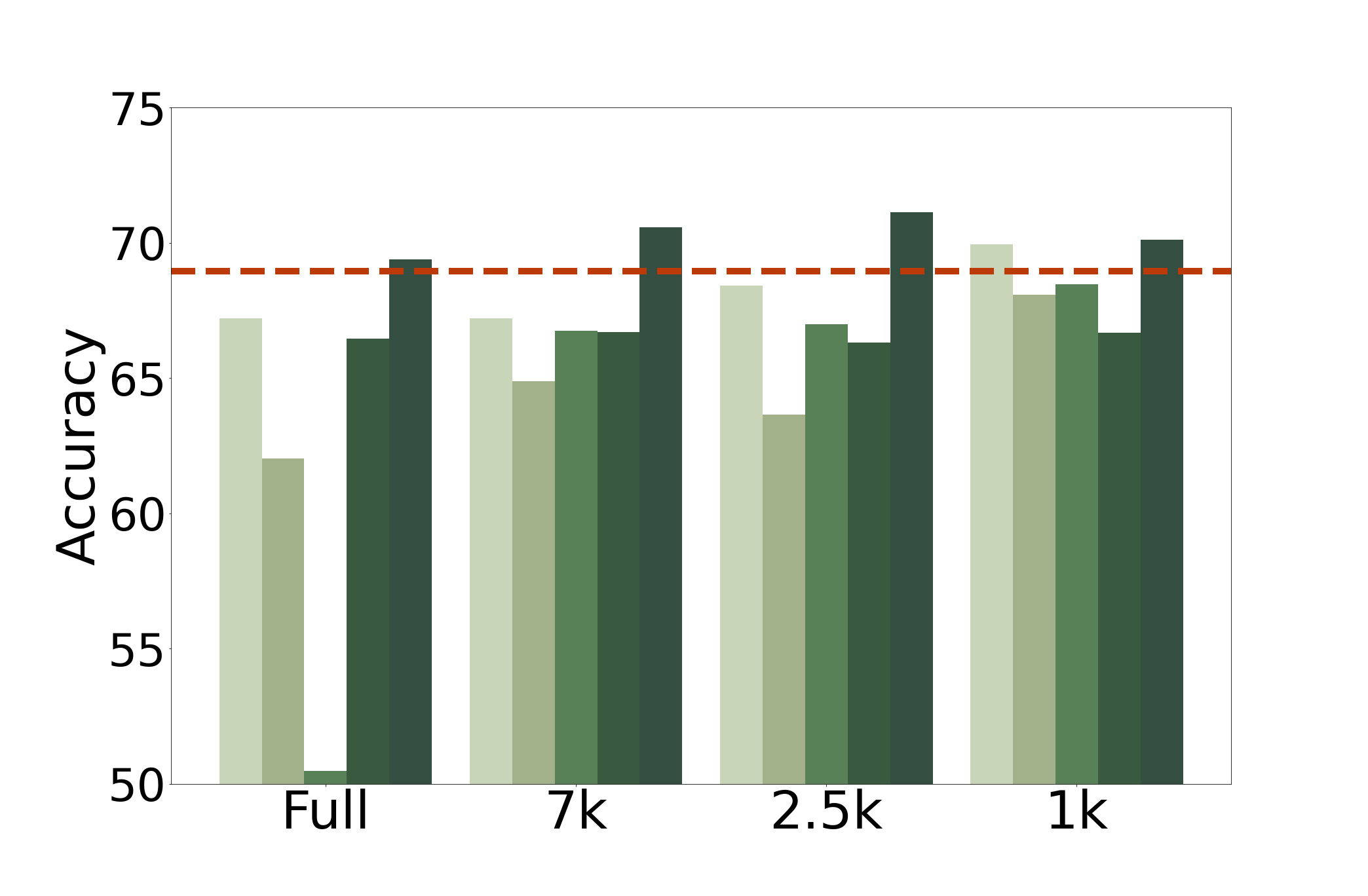}\label{fig:datasize-coin}}
    \subfigure[Semantic Odd Man Out]{\includegraphics[width=7cm,height=5cm]{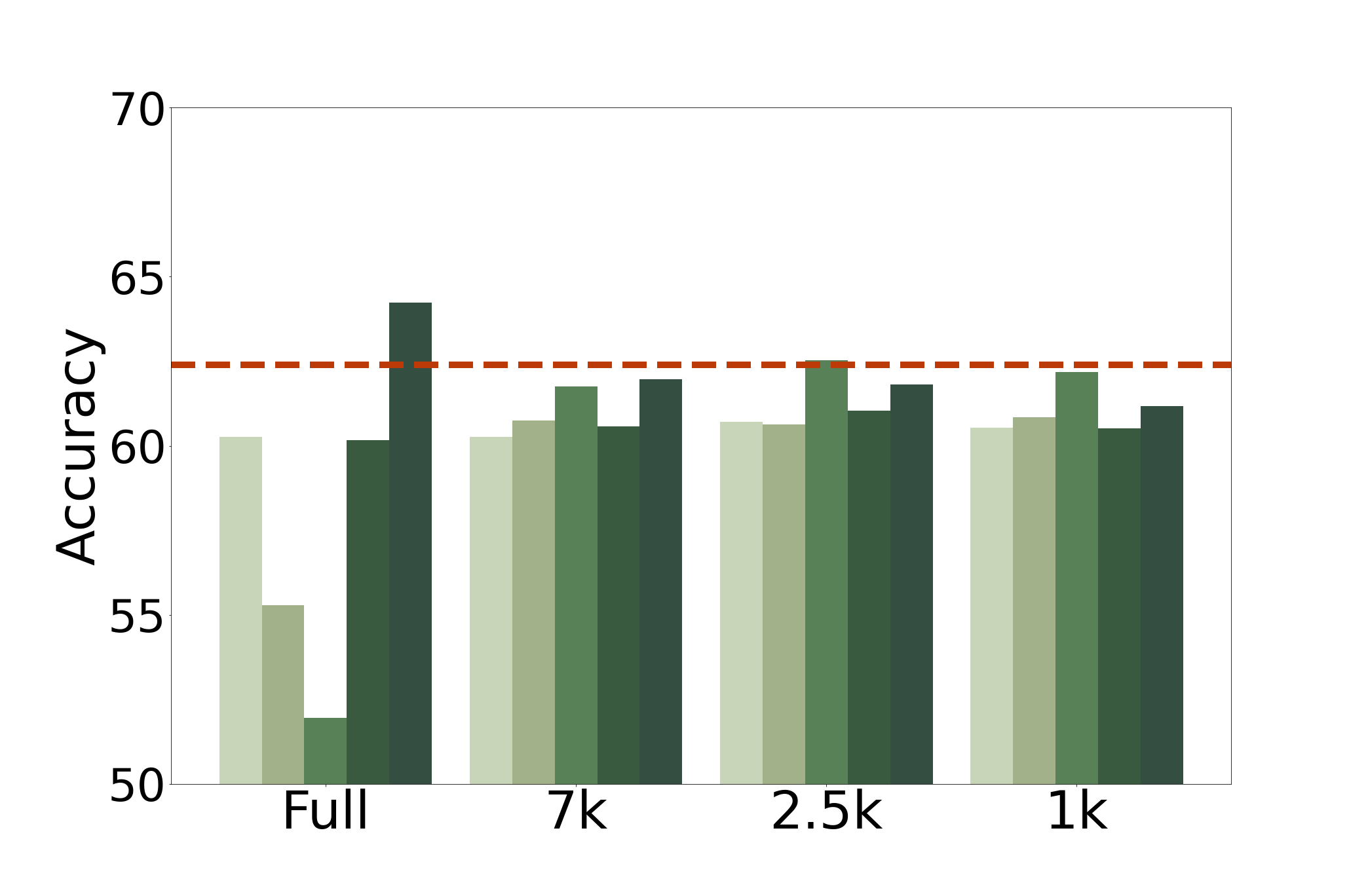} \label{fig:datasize-omo}} 

    \caption{An illustration of the probing performance of models fine-tuned on fixed-size training sets of five different tasks. The pre-trained BERT’s performance on each of the four probing tasks has been shown by the dashed red line. The figures suggest that different fine-tuned models, irrespective of the fine-tuning task, almost encode similar linguistic knowledge when trained on equal-sized data.}
    \label{fig:fig-datasize}
    
\end{figure*}


\section{Data Size Analysis}
\label{sec:datasize-analysis}
In this section, we first provide insight on the role of target tasks in capturing or forgetting different types of knowledge (e.g., syntactic and semantic) during fine-tuning. Then, we investigate the role of datasets' size on linguistic knowledge.

\subsection{Probing Linguistic Knowledge}
\label{sec:Probing}
We empirically evaluate the linguistic knowledge captured by several fine-tuned models through the lens of probing performance.
Figure \ref{fig:fig-probe-ling} illustrates the layer-wise probing performance of fine-tuned models, considering pre-trained BERT as our baseline. As can be observed, different models carry similar linguistic knowledge up to the middle layers, and the difference gradually increases as we move up to the higher layers. This observation is consistent with the reported results by \citet{merchant-etal-2020-happens}. Their experimental analysis indicates that fine-tuning mostly changes the higher layers while having a smaller impact on the lower layers. \citet{durrani-etal-2021-transfer} also reported a similar behavior in other LMs through different probing tasks.

The results illustrated in Figure~\ref{fig:fig-probe-ling} clearly highlight the impact of data size on probing accuracy.
We can observe that the probing performance of the baseline and models fine-tuned on smaller datasets {(e.g., MRPC, SST-2, and CoLA)} are comparable, whereas fine-tuning on larger data sizes (e.g., QQP and MNLI) seems to have impacted probing performance by a significant margin. 
In what follows, we carry out experiments to better understand the reasons behind this observation.

\subsection{The Impact of Data Size}
One of the popular studies in probing is investigating the changes made to a model's linguistic knowledge after fine-tuning.
The changes brought about in the model upon fine-tuning are taken as a means to explain the nature of the corresponding task on which fine-tuning has been carried out \citep{durrani-etal-2021-transfer}.
Existing studies usually consider several tasks, many of which do not have datasets of comparable size.
For instance, {in the GLUE benchmark,} MNLI is 46 times larger than CoLA.
These studies usually focus on the type of downstream tasks only, overlooking the size of their datasets.

Based on our observations in Section~\ref{sec:Probing}, we hypothesize that, in addition to the type of the downstream task, the size of its corresponding dataset can play an important role in improving or impairing the linguistic knowledge encoded in the model. 
We examined our hypothesis by fine-tuning pre-trained BERT on the selected downstream tasks with different sets of samples.
Specifically, taking the pre-trained BERT as the baseline, we analyze the effect of the training set size on the encoded linguistic knowledge by limiting the number of samples to 7k, 2.5k, and 1k. 
Figure \ref{fig:fig-datasize} shows the results of this experiment. 
In general, the results confirm our hypothesis that data size plays a significant role in probing accuracy. 
In what follows, we further discuss our observations from this experiment.

\begin{table*}[t!]
\resizebox{\textwidth}{!}{%
\setlength{\tabcolsep}{12pt}
\begin{tabular}{@{}clcccccccccccc@{}}

\toprule
&&
\multicolumn{1}{l}{}   &           \multicolumn{5}{c}{Bigram Shift}                                                             &                                                                                                                                               & \multicolumn{5}{c}{Semantic Odd Man Out} \\ 
\cmidrule(lr){3-8}
\cmidrule(l){10-14}

\multicolumn{1}{l}{}   &          & & \textbf{Full}                                         & \textbf{7k}                                         & \textbf{2.5k}                                       & \textbf{1k}                                         & \textbf{baseline} &&  \textbf{Full}                                         & \textbf{7k}                                         & \textbf{2.5k}                                       & \textbf{1k}                                         & \textbf{baseline} \\ \cmidrule(l){2-14}
                       & Layer 2  && \cellcolor[HTML]{FFFCFB}-0.49~                         & \cellcolor[HTML]{FAFCFC}0.16                        & \cellcolor[HTML]{FFFBFA}-0.63~                       & \cellcolor[HTML]{FFFAF8}-0.82~                       & 53.60            & & \cellcolor[HTML]{FFF4F1}-0.65                         & \cellcolor[HTML]{FFFAF9}-0.25~                       & \cellcolor[HTML]{FFFDFD}-0.06~                       & \cellcolor[HTML]{FFFBFA}-0.23~                       & 53.92             \\
                       & Layer 7  && \cellcolor[HTML]{C5DEDE}{\color[HTML]{000000} 1.78}   & \cellcolor[HTML]{D3E6E6}1.36                        & \cellcolor[HTML]{CCE2E2}1.57                        & \cellcolor[HTML]{BDD9D9}2.03                        & 75.93            & & \cellcolor[HTML]{FFC6B9}-3.40~                         & \cellcolor[HTML]{FFD8CF}-2.31~                       & \cellcolor[HTML]{FFF1EE}-0.80~                       & \cellcolor[HTML]{FFE7E1}-1.43~                       & 59.41             \\
                       & Layer 11 && \cellcolor[HTML]{207E7F}{\color[HTML]{FFFFFF} 6.78}   & \cellcolor[HTML]{157879}{\color[HTML]{FFFFFF} 7.09} & \cellcolor[HTML]{308889}{\color[HTML]{FFFFFF} 6.29} & \cellcolor[HTML]{579E9F}{\color[HTML]{FFFFFF} 5.10} & 82.39            & & \cellcolor[HTML]{157879}{\color[HTML]{FFFFFF} 2.08}   & \cellcolor[HTML]{378C8D}{\color[HTML]{FFFFFF} 1.78} & \cellcolor[HTML]{32898A}{\color[HTML]{FFFFFF} 1.83} & \cellcolor[HTML]{91C0C0}0.98                        & 61.32             \\
\multirow{-4}{*}{\rotatebox{90}{CoLA}} & Layer 12 && \cellcolor[HTML]{32898A}{\color[HTML]{FFFFFF} 6.22}   & \cellcolor[HTML]{378C8C}{\color[HTML]{FFFFFF} 6.09} & \cellcolor[HTML]{489697}{\color[HTML]{FFFFFF} 5.56} & \cellcolor[HTML]{5FA3A4}{\color[HTML]{FFFFFF} 4.85} & 83.23    &         & \cellcolor[HTML]{308889}{\color[HTML]{FFFFFF} 1.84}   & \cellcolor[HTML]{FFF7F5}-0.44~                       & \cellcolor[HTML]{FFF5F3}-0.58~                       & \cellcolor[HTML]{FFEAE5}-1.23~                       & 62.40             \\ \cmidrule(l){2-14} 
                       & Layer 2 & & \cellcolor[HTML]{FFFAF9}-0.74~                         & \cellcolor[HTML]{FFFAF8}-0.82~                       & \cellcolor[HTML]{FFFDFC}-0.30~                       & \cellcolor[HTML]{FFF9F8}-0.94~                       & 53.60        &     & \cellcolor[HTML]{FFF5F3}-0.55~                         & \cellcolor[HTML]{FFF5F3}-0.55~                       & \cellcolor[HTML]{FFF6F4}-0.52~                       & \cellcolor[HTML]{FFFDFC}-0.10~~                       & 53.92             \\
                       & Layer 7  && \cellcolor[HTML]{FFF1EE}-2.26~                         & \cellcolor[HTML]{FFF3F0}-1.94~                       & \cellcolor[HTML]{FFF3F0}-1.94~                       & \cellcolor[HTML]{FFFDFD}-0.24~                       & 75.93         &    & \cellcolor[HTML]{FFE0D9}-1.81~                         & \cellcolor[HTML]{FFE4DF}-1.56~                       & \cellcolor[HTML]{FFE9E4}-1.29~                       & \cellcolor[HTML]{FFEAE6}-1.22~                       & 59.41             \\
                       & Layer 11 && \cellcolor[HTML]{FFE8E3}-3.81~                         & \cellcolor[HTML]{FFF0EC}-2.48~                       & \cellcolor[HTML]{FFF3F1}-1.89~                       & \cellcolor[HTML]{FFF7F5}-1.33~                       & 82.39          &   & \cellcolor[HTML]{FFE8E3}-1.33~                         & \cellcolor[HTML]{FFF0ED}-0.87~                       & \cellcolor[HTML]{FFF0EC}-0.88~                       & \cellcolor[HTML]{FFF5F3}-0.55~                       & 61.32             \\
\multirow{-4}{*}{\rotatebox{90}{SST-2}} & Layer 12 && \cellcolor[HTML]{FFDCD4}-5.77~                         & \cellcolor[HTML]{FFE1DB}-4.87~                       & \cellcolor[HTML]{FFEAE6}-3.40~                       & \cellcolor[HTML]{FFEBE7}-3.20~                       & 83.23        &     & \cellcolor[HTML]{FFD9D1}-2.24~                         & \cellcolor[HTML]{FFE0D9}-1.83~                       & \cellcolor[HTML]{FFE8E2}-1.37~                       & \cellcolor[HTML]{FFDFD8}-1.89~                       & 62.40             \\ \cmidrule(l){2-14} 
                       & Layer 2 & & \cellcolor[HTML]{FFF2F0}-2.01~                         & \cellcolor[HTML]{FFFAF9}-0.78~                       & \cellcolor[HTML]{FFFDFC}-0.32~                       & \cellcolor[HTML]{EFF6F6}0.51                        & 53.60          &   & \cellcolor[HTML]{FFE2DC}-1.69~                         & \cellcolor[HTML]{FFF8F7}-0.38~                       & \cellcolor[HTML]{FFF4F2}-0.62~                       & \cellcolor[HTML]{FFFCFC}-0.13~                       & 53.92             \\
                       & Layer 7  && \cellcolor[HTML]{FFCFC4}-7.94~                         & \cellcolor[HTML]{FFF4F2}-1.68~                       & \cellcolor[HTML]{FFF9F8}-0.85~                       & \cellcolor[HTML]{FFFAF8}-0.83~                       & 75.93           &  & \cellcolor[HTML]{FFD4CA}-2.55~                         & \cellcolor[HTML]{FFF5F3}-0.54~                       & \cellcolor[HTML]{FFF2EF}-0.74~                       & \cellcolor[HTML]{FFD3C9}-2.61~                       & 59.41             \\
                       & Layer 11 && \cellcolor[HTML]{FF967F}{\color[HTML]{000000} -17.31~~~} & \cellcolor[HTML]{FFD7CE}-6.54~                       & \cellcolor[HTML]{FFE4DE}-4.49~                       & \cellcolor[HTML]{FFF5F3}-1.52~                       & 82.39        &     & \cellcolor[HTML]{FFA793}-5.25~                         & \cellcolor[HTML]{FFF9F8}-0.32~                       & \cellcolor[HTML]{FFE9E4}-1.30~                       & \cellcolor[HTML]{FFF7F5}-0.45~                       & 61.32             \\
\multirow{-4}{*}{\rotatebox{90}{MNLI}} & Layer 12 & & \cellcolor[HTML]{FF896F}{\color[HTML]{000000} -19.52~~~} & \cellcolor[HTML]{FFC9BE}-8.84~                       & \cellcolor[HTML]{FFD8CF}-6.44~                       & \cellcolor[HTML]{FFECE7}-3.14~                       & 83.23       &      & \cellcolor[HTML]{FF876D}{\color[HTML]{000000} -7.12~}  & \cellcolor[HTML]{FFE3DD}-1.65~                       & \cellcolor[HTML]{FFE1DA}-1.76~                       & \cellcolor[HTML]{FFE5DF}-1.55~                       & 62.40             \\ \cmidrule(l){2-14} 
                       & Layer 2  && \cellcolor[HTML]{C0DBDB}1.93                          & \cellcolor[HTML]{E9F3F3}0.68                        & \cellcolor[HTML]{F4F9F9}0.35                        & \cellcolor[HTML]{FFFDFD}-0.26~                       & 53.60       &      & \cellcolor[HTML]{FFF7F5}-0.46~                         & \cellcolor[HTML]{FFFCFC}-0.12~                       & \cellcolor[HTML]{FFFAF9}-0.27~                       & \cellcolor[HTML]{FFFBFA}-0.21~                       & 53.92             \\
                       & Layer 7  && \cellcolor[HTML]{FFB3A2}-12.63~~~                        & \cellcolor[HTML]{FFF5F3}-1.55~                       & \cellcolor[HTML]{FFFEFE}-0.05~                       & \cellcolor[HTML]{ECF4F4}0.60                        & 75.93        &     & \cellcolor[HTML]{FFAE9C}-4.82~                         & \cellcolor[HTML]{FFFEFE}-0.01~                       & \cellcolor[HTML]{DEECEC}0.30                        & \cellcolor[HTML]{FFF6F4}-0.53~                       & 59.41             \\
                       & Layer 11 & &\cellcolor[HTML]{FF5C38}{\color[HTML]{FFFFFF} -26.97~~~} & \cellcolor[HTML]{FFE8E3}-3.78~                       & \cellcolor[HTML]{FFF8F7}-1.05~                       & \cellcolor[HTML]{FFF0EC}-2.46~                       & 82.39         &    & \cellcolor[HTML]{FF6442}{\color[HTML]{FFFFFF} -9.22~}  & \cellcolor[HTML]{9BC6C6}0.89                        & \cellcolor[HTML]{9AC5C6}0.90                        & \cellcolor[HTML]{B6D5D6}0.65                        & 61.32             \\
\multirow{-4}{*}{\rotatebox{90}{QQP}}  & Layer 12 && \cellcolor[HTML]{FF5029}{\color[HTML]{FFFFFF} -29.12~~~} & \cellcolor[HTML]{FFDCD5}-5.70~                       & \cellcolor[HTML]{FFF4F1}-1.81~                       & \cellcolor[HTML]{FFECE8}-3.00~                       & 83.23        &     & \cellcolor[HTML]{FF5029}{\color[HTML]{FFFFFF} -10.45~~~} & \cellcolor[HTML]{FFF4F1}-0.65~                       & \cellcolor[HTML]{F1F7F7}0.13                        & \cellcolor[HTML]{FFFBFA}-0.22~                       & 62.40             \\ \cmidrule(l){2-14} 
                       & Layer 2  && \cellcolor[HTML]{FFF8F7}-1.08~                         &  \textemdash                            & \cellcolor[HTML]{FFFAF8}-0.82~                       & \cellcolor[HTML]{FFF9F7}-0.96~                       & 53.60          &   & \cellcolor[HTML]{FFF8F7}-0.37~                         & \textemdash                          & \cellcolor[HTML]{FFF5F3}-0.56~                       & \cellcolor[HTML]{FFF6F4}-0.53~                       & 53.92             \\
                       & Layer 7  && \cellcolor[HTML]{FFFBFB}-0.53~                         &     \textemdash                       & \cellcolor[HTML]{FFF8F7}-1.04~                       & \cellcolor[HTML]{FFFEFE}-0.09~                       & 75.93            & & \cellcolor[HTML]{FFF8F7}-0.36~                         & \textemdash                            & \cellcolor[HTML]{DFEDED}0.29                        & \cellcolor[HTML]{FFF9F8}-0.34~                       & 59.41             \\
                       & Layer 11 && \cellcolor[HTML]{FFF3F0}-1.94~                         & \textemdash                & \cellcolor[HTML]{FFF3F1}-1.90~                        & \cellcolor[HTML]{FFF6F4}-1.41~                       & 82.39             & & \cellcolor[HTML]{FFEDE9}-1.05~                         & \textemdash                           & \cellcolor[HTML]{67A7A8}{\color[HTML]{000000} 1.36} & \cellcolor[HTML]{68A8A9}{\color[HTML]{000000} 1.35} & 61.32             \\
\multirow{-4}{*}{\rotatebox{90}{MRPC}} & Layer 12 && \cellcolor[HTML]{FFE7E2}-3.87~                         & \textemdash                          & \cellcolor[HTML]{FFEAE5}-3.45~                       & \cellcolor[HTML]{FFF1EE}-2.31~                       & 83.23        &     & \cellcolor[HTML]{FFDBD3}-2.13~                         & \textemdash                           & \cellcolor[HTML]{FFE2DC}-1.70~                        & \cellcolor[HTML]{FFDFD8}-1.86~                       & 62.40             \\ \bottomrule

\end{tabular}%
}
\caption{Layer-wise performance of models on the probing tasks. Each cell represents the difference (delta) in performance between the corresponding fine-tuned model and the baseline. The pre-trained BERT performance (baseline) is shown in the right columns.
}
\label{tab:layer-wise}
\end{table*}

\subsection{Discussion}
The effect of data size on both syntactic and semantic probing tasks is notable, denoted by the large gaps between the probing results of the models fine-tuned on larger data sizes and the baseline (see Figure~\ref{fig:fig-probe-ling}).
We observe that as the number of samples increases, the gap between fine-tuned models and the pre-trained BERT (baseline) becomes more apparent. For instance, probing the model fine-tuned on QQP's full training set demonstrates that it has far less linguistic knowledge than the baseline. However, after fine-tuning the model on QQP with fewer training samples (7k, 2.5, and 1k), {not much change is observed across the results}. This shows that fine-tuning data size indeed affects the linguistic knowledge encoded by the model. 

Overall, we can conclude that the amount of linguistic knowledge through fine-tuning is highly affected by data size. 
This suggests that data size should be taken into account when analyzing fine-tuned models.


\section{Layer-wise Analysis}
\label{sec:layer-wise}

Given our observations on the role of data size, we were curious to see {how it affects the encoded knowledge in specific layers.}
As noted by \citet{jawahar-etal-2019-bert}, BERT's layers can be divided into three classes in terms of the linguistic knowledge they capture. 
To this end, we carry out experiments by probing layers 2, 7, and 11-12 to cover all the three categories.

Table \ref{tab:layer-wise} shows our results obtained from this experiment, which are compared with BERT-base. Due to our limited resources and the excessive number of experiments, we omitted probing tasks that did not show any distinguishable patterns (Figures \ref{fig:fig-probe-ling} and \ref{fig:fig-datasize}), i.e., 
Coordination Inversion and Object Number.
The results follow a similar trend to the ones depicted in Figure~\ref{fig:fig-datasize}. As we decrease the number of training samples, the probing performance on the fine-tuned models gets closer to the baseline across all layers. MNLI and QQP's behaviors are compelling evidence of the effectiveness of data size across layers. Such models fine-tuned on larger datasets undergo more considerable changes than those with smaller data sizes.

Regardless of data size, we can also observe that fine-tuning mainly affects higher layers. Our finding is aligned with \citet{merchant-etal-2020-happens} that fine-tuning has a more significant impact on higher layers and negligible effects on lower layers. 
There is also an interesting pattern concerning CoLA's performance. Despite a drop in performance of around 15\% from the full to 1k version (Table~\ref{statistics-table}), the linguistic knowledge has been marginally affected by data size. We leave further investigations on this to future work.


\begin{table}[t!]
\scalebox{0.80}{
\centering
\setlength{\tabcolsep}{14pt}
\begin{tabular}{@{}clcccl@{}}
\toprule
\multicolumn{1}{l}{}   &          & \textbf{Full} & \textbf{7k}                   & \textbf{2.5k}                 \\ 
\midrule
\multicolumn{1}{l}{}   & \multicolumn{4}{c}{Bigram Shift}                                                         \\ \cmidrule(l){2-5} 
                       & Layer 2  & 52.87        & \cellcolor[HTML]{FEFFFF}0.07  & \cellcolor[HTML]{FEFEFE}-0.03 \\
                       & Layer 7  & 71.88        & \cellcolor[HTML]{E6C9CA}-2.08~ & \cellcolor[HTML]{F2E2E2}-1.12~ \\
                       & Layer 11 & 74.08        & \cellcolor[HTML]{F8FBFC}0.49  & \cellcolor[HTML]{D3E2EC}2.90  \\
\multirow{-4}{*}{\rotatebox{90}{QQP}}  & Layer 12 & 73.25        & \cellcolor[HTML]{FDFCFC}-0.10~ & \cellcolor[HTML]{E4EDF3}1.81  \\ \cmidrule(l){2-5} 
                       & Layer 2  & 51.9         & \cellcolor[HTML]{FCF8F8}-0.24~ & \cellcolor[HTML]{F1E1E1}-1.16~ \\
                       & Layer 7  & 71.03        & \cellcolor[HTML]{F2F7FA}0.88  & \cellcolor[HTML]{FEFEFE}-0.02~ \\
                       & Layer 11 & 67.69        & \cellcolor[HTML]{E2ECF3}1.93  & \cellcolor[HTML]{DAE7EF}2.47  \\
\multirow{-4}{*}{\rotatebox{90}{MNLI}} & Layer 12 & 65.82        & \cellcolor[HTML]{E9F1F6}1.48  & \cellcolor[HTML]{E7F0F5}1.57  \\ \midrule
\multicolumn{1}{l}{}   & \multicolumn{4}{c}{Semantic Odd Man Out}                                                 \\ \cmidrule(l){2-5}
                       & Layer 2  & 53.73        & \cellcolor[HTML]{F4F8FB}0.73  & \cellcolor[HTML]{F8FBFC}0.49  \\
                       & Layer 7  & 56.12        & \cellcolor[HTML]{F1F6F9}0.95  & \cellcolor[HTML]{E7EFF5}1.61  \\
                       & Layer 11 & 58.11        & \cellcolor[HTML]{EDF3F7}1.23  & \cellcolor[HTML]{EEF4F8}1.16  \\
\multirow{-4}{*}{\rotatebox{90}{QQP}}  & Layer 12 & 58.03        & \cellcolor[HTML]{EBF2F7}1.34  & \cellcolor[HTML]{FBFCFD}0.31  \\ \cmidrule(l){2-5} 
                       & Layer 2  & 53.23        & \cellcolor[HTML]{FCFDFE}0.24  & \cellcolor[HTML]{F4F8FA}0.76  \\
                       & Layer 7  & 57.00          & \cellcolor[HTML]{E8F0F5}1.54  & \cellcolor[HTML]{E7F0F5}1.60  \\
                       & Layer 11 & 57.27        & \cellcolor[HTML]{DFEBF1}2.10  & \cellcolor[HTML]{EEF4F8}1.17  \\
\multirow{-4}{*}{\rotatebox{90}{MNLI}} & Layer 12 & 56.77        & \cellcolor[HTML]{DAE7EF}2.43  & \cellcolor[HTML]{EDF3F7}1.22  \\ 
\bottomrule
\end{tabular}%
}
\caption{The performance of models trained with fixed and equal number of iterations across different sizes on each downstream task. Every cell demonstrates the difference (delta) between the full and the fixed-sized models. With an equal number of iterations, in each layer, fine-tuned models have a similar performance.}
\label{tab:fixed-iter}
\end{table}
 
\section{Fixed Iteration Analysis}

Given the observations from Section~\ref{sec:layer-wise}, we have realized that by training BERT on larger datasets, the model's performance deviates substantially from the baseline. However, by reducing the size of training data, the gap between the fine-tuned models and the baseline decreases. This behavior can be either attributed to the diversity of training samples or to the larger number of iterations through which the model is updated. 

To address this, we repeated the same experiments carried out in Section~\ref{sec:layer-wise} but with fixing the number of iterations on all data sizes. This allows the model to be fine-tuned for an equal number of iterations across different data sizes of a specific task.
Note that we fine-tuned the full models for just one epoch to avoid a large number of iterations for the 7k and 2.5k models.
Since SST-2, CoLA, and MRPC have much smaller datasets, and the number of iterations does not substantially differ across the full, 7k, and 2.5k models, we have dropped them from this scenario.

Table \ref{tab:fixed-iter} summarizes our results. The first interesting pattern is that fine-tuning for more epochs significantly impairs the captured linguistic knowledge. For instance, we can observe the impact of longer training by comparing Bigram Shift performance on QQP across Tables \ref{tab:layer-wise} (54.11) and \ref{tab:fixed-iter} (73.25)\footnote{As mentioned in Section~\ref{sec:fine-tuning-setup}, the models in Table~\ref{tab:layer-wise} were fine-tuned for five epochs.}.
As Table \ref{tab:fixed-iter} suggests, fixing the number of iterations reduces the gap across different data sizes, making the 7k and 2.5k models behave almost similarly to the full models.
For instance, in Table \ref{tab:layer-wise}, there is a gap of 24\% in the last layer's performance between the full and the 7k QQP on Bigram Shift, which has been reduced to approximately $-$0.1 with equal training steps (Table \ref{tab:layer-wise}).

This finding is interesting because, firstly, it indicates that the high variance between baselines and full models is mainly due to the number of times their weights are updated during fine-tuning rather than the diversity of the training samples. Secondly, with equal data sizes, the role of target tasks becomes less influential in the linguistic knowledge introduced into the model by fine-tuning, reinforcing the conclusions from Section~\ref{sec:layer-wise}.

\begin{figure*}
    \centering
    \includegraphics[width = 3.89cm]{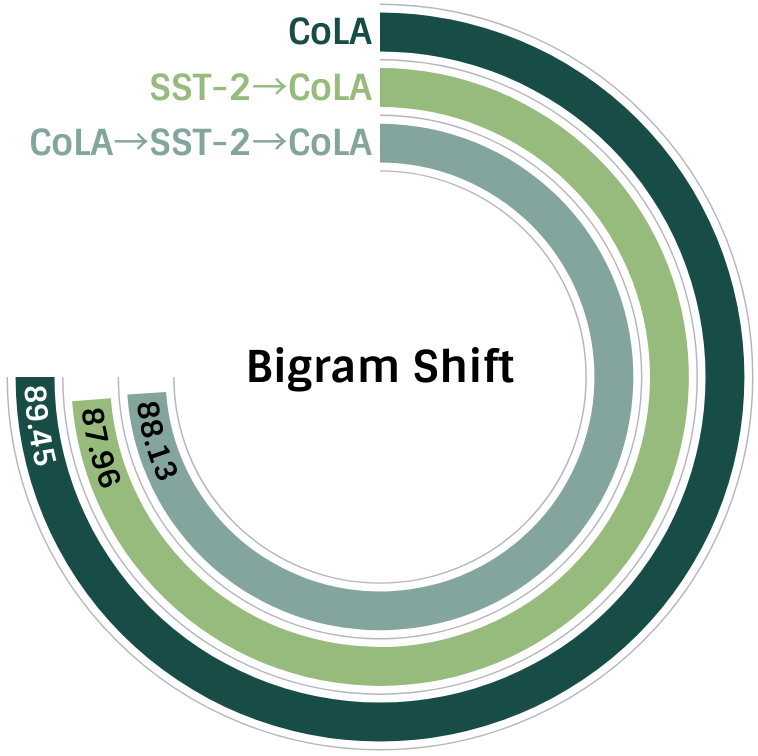}
    \includegraphics[width = 3.89cm]{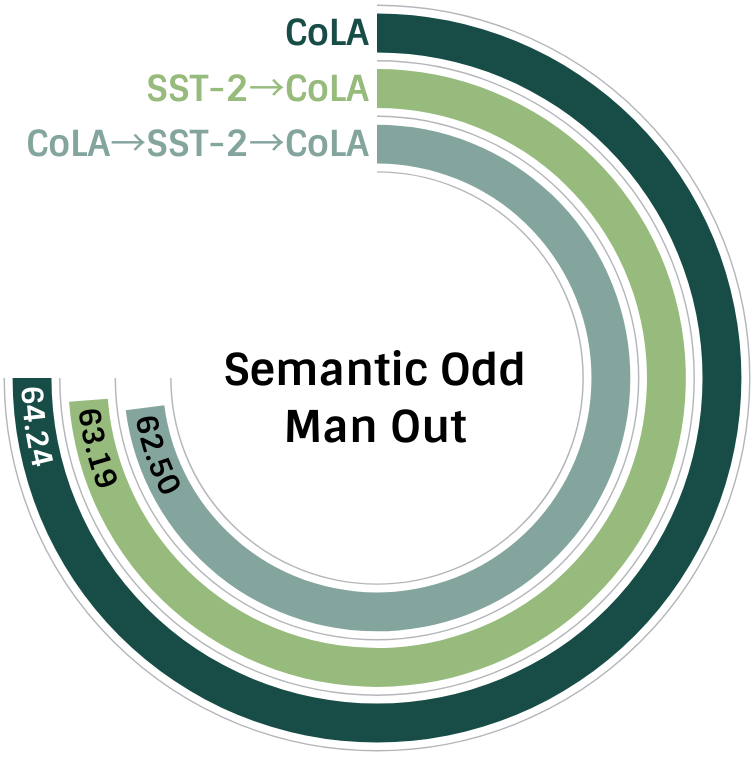}
    \includegraphics[width = 3.89cm]{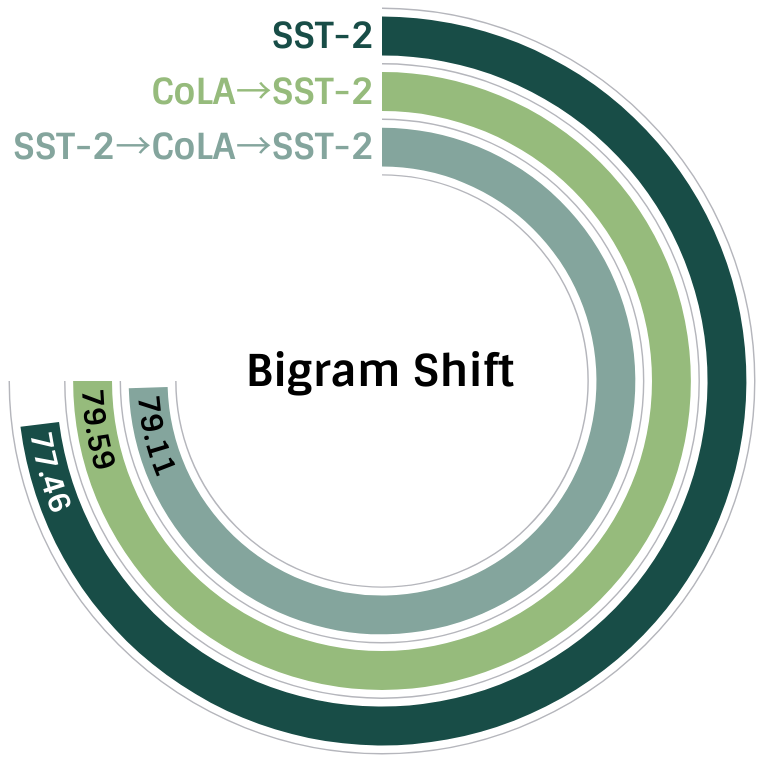}
    \includegraphics[width = 3.89cm]{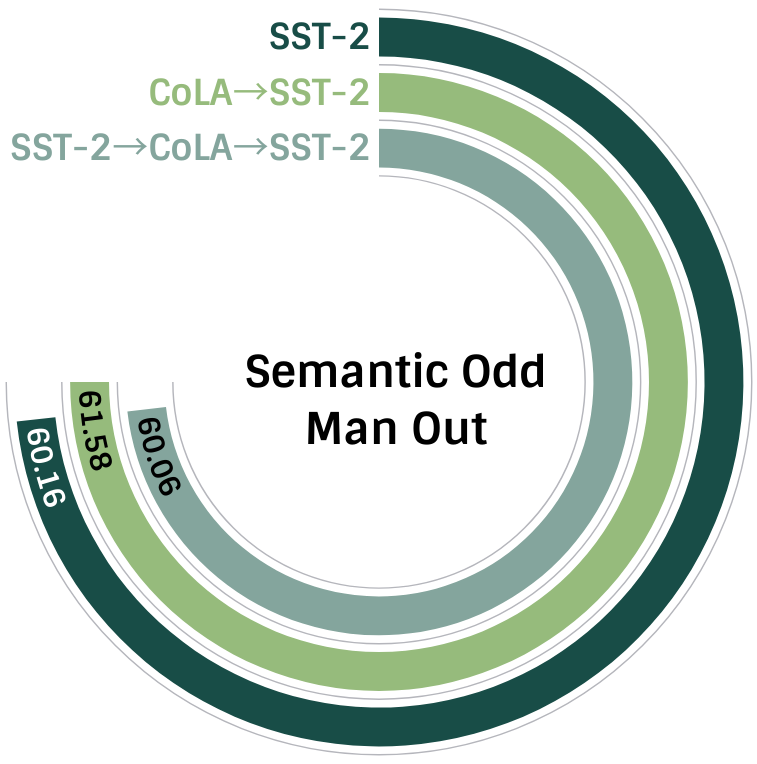}
    \includegraphics[width = 3.9cm]{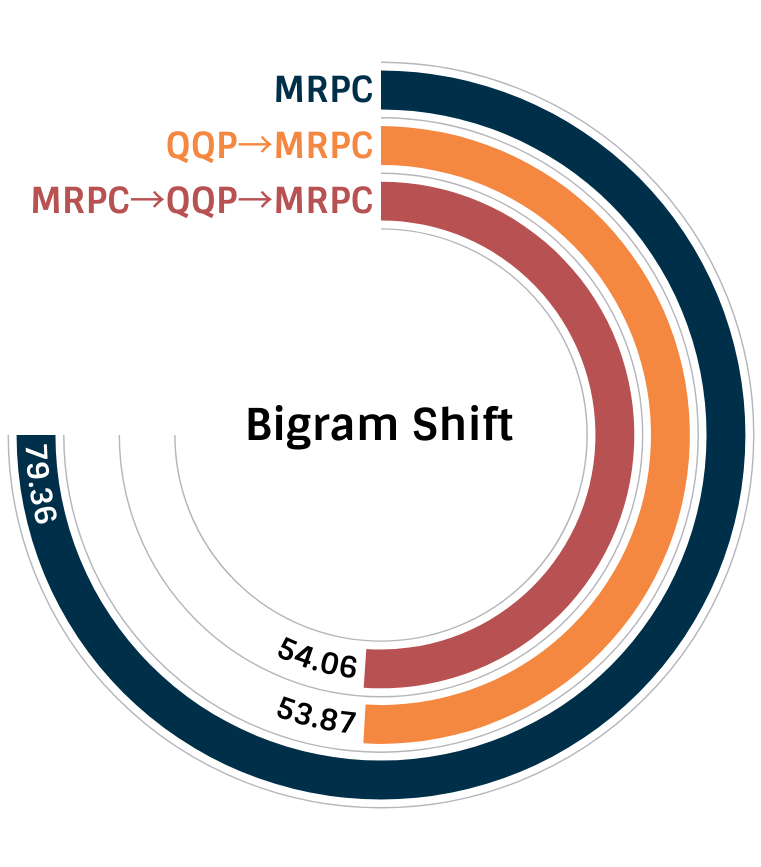}
    \includegraphics[width = 3.9cm]{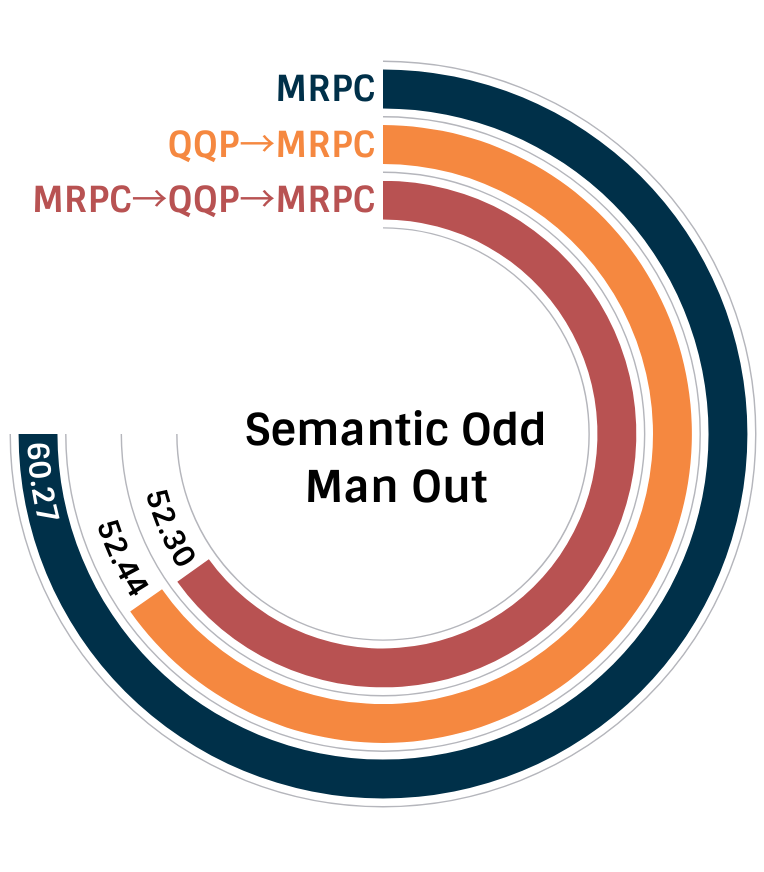}
    \includegraphics[width = 3.9cm]{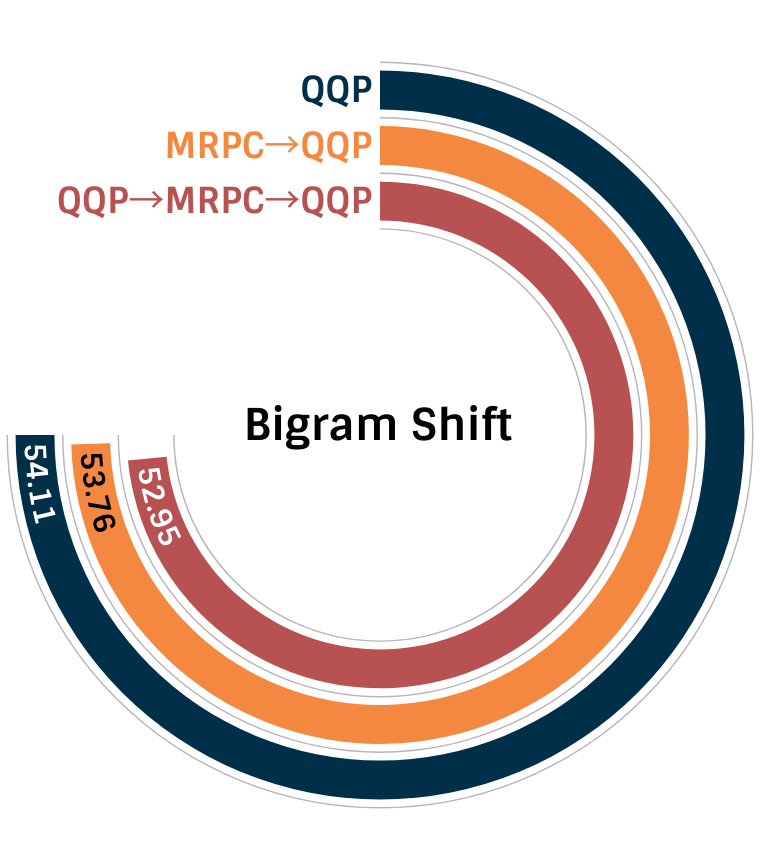}
    \includegraphics[width = 3.9cm]{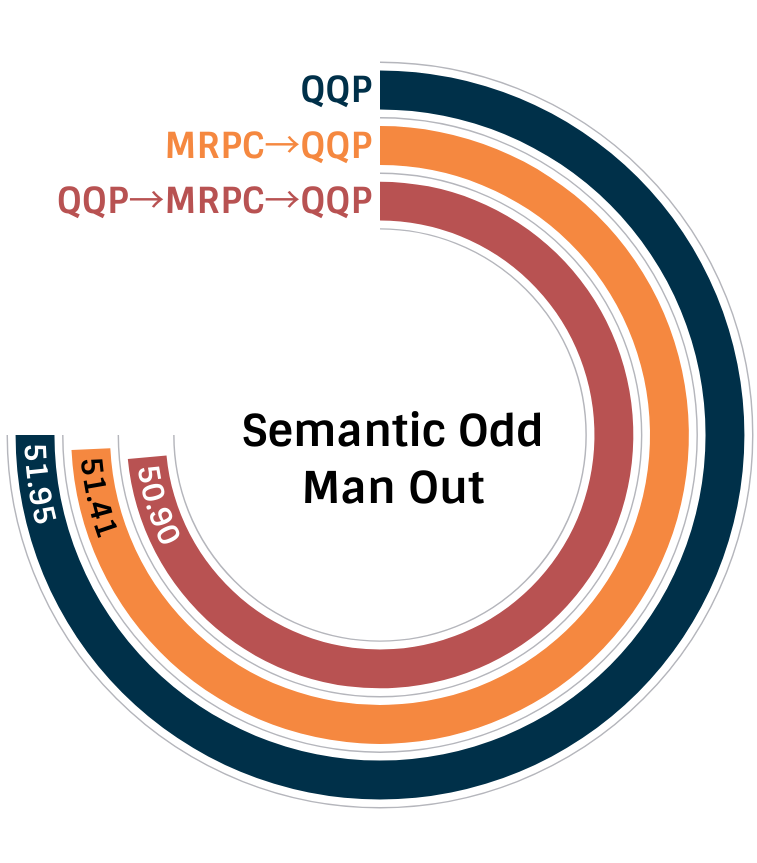}
    \caption{The performance of the models after being sequentially fine-tuned on different tasks. Three-quarters of a circle represents the maximum value and the outer circle is the baseline. The figures demonstrate that the modified knowledge recoverability depends on the fine-tuning data size.}
    \label{fig:sequence-analysis}
\end{figure*}

\section{{Linguistic Knowledge Recoverability}}
Fine-tuning procedure modifies the encoded linguistic knowledge in the pre-trained model. 
In this section, we aim at verifying the extent to which these modifications are recoverable. 
To this end, taking a fine-tuned model on a specific task as our baseline, we further fine-tune the model on another task. We then compare the probing performance of the resulting models with their corresponding baselines. 
High similarity in probing performance indicates the recoverability of the modifications.

We opt for CoLA and SST-2 as a pair of tasks with different linguistic objectives but with comparable data sizes.
Also, we experiment with MRPC and QQP, which are similar tasks but with significantly different data sizes (the former’s data size is a hundred times larger than the latter’s). 
For instance, considering CoLA and SST-2 as our fine-tuning task pair, \textit{SST-2 $\rightarrow$ CoLA $\rightarrow$ SST-2} stands for a setting where we have consecutively fine-tuned the model on SST-2, CoLA, and SST-2. Following our previous experiments, we report the probing results for the Bigram Shift and Semantic Odd Man Out tasks.\footnote{More results for the Object Number and Coordination Inversion tasks can be found in Appendix~\ref{sec:appendixB}.}

The results are presented in Figure~\ref{fig:sequence-analysis}. 
The three-quarters of a circle in the figures represent the maximum value in the {corresponding} probing task. As shown in the figures, the linguistic knowledge is recoverable through re-fine-tuning on a set of pairs with comparable data sizes. In the previous sections, we observed that CoLA and SST-2 have notably different performances on Bigram Shift and Semantic Odd Man Out. Nevertheless, after re-fine-tuning, both target tasks can recover the knowledge modified by the previous fine-tuning step.

On the other hand, for the QQP and MRPC pair, we observe a different behavior in which the data size of QQP highly limits the extent of knowledge recoverability. Considering Bigram Shift, we observe that the final MRPC fine-tuning in the \textit{QQP $\rightarrow$ MRPC} and \textit{MRPC $\rightarrow$ QQP $\rightarrow$ MRPC} settings can not recover the modification introduced by QQP (the probing results remain similar to QQP’s). In the reverse setting (\textit{MRPC $\rightarrow$ QQP} and \textit{QQP $\rightarrow$ MRPC $\rightarrow$ QQP}), the probing performance is negligently affected by MRPC data size, leading to a performance fairly similar to QQP’s.\footnote{We have also carried out the exact experiments with QQP 7k to make sure the results are related to the size of the tasks.}

Our results suggest that the extent of knowledge recoverability is bound to the fine-tuning data size. {More specifically, further fine-tuning a fine-tuned model with a comparable data size (e.g., \textit{SST-2 $\rightarrow$ CoLA} and \textit{CoLA $\rightarrow$ SST-2 $\rightarrow$ CoLA} introduces the same modifications as fine-tuning a pre-trained model (e.g., CoLA).} 
However, increasing the data size in one of these tasks decreases the extent of recoverability by the other task.

\section{Conclusion}
In this paper, we carried out a set of experiments to determine the effect of training data size on the probing performance of fine-tuned models. To begin with, by individually probing all layers, we found out that models fine-tuned on larger datasets deviate more from the base model in terms of their encoded linguistic knowledge. Therefore, we argue that comparing the linguistic knowledge of fine-tuned models is valid only if they are trained on datasets of comparable sizes.
Through layer-wise probing analysis, we realized that the number of training samples mainly affects the probing results for the higher layers, while the results remain similar in the lower layers across different target tasks. Furthermore, we investigated why data size affects the probing performance of fine-tuned models through training the models with limited training data for the same number of iterations as we trained the full models. We showed that the gap in probing performance between models fine-tuned on different data sizes is due to the number of iterations for which the model is updated during fine-tuning rather than the diversity of the training set.
Finally, in our last experiment, we showed that the size of a target task's dataset affects the extent to which it can recover the linguistic knowledge previously changed by a different task.

We argue that probing accuracy cannot fully represent the linguistic knowledge captured by fine-tuned models, given that factors, such as the size of the dataset, can highly affect probing accuracy and should be ruled out in any such study. 
As future work, we plan to evaluate the reliability of existing accuracy and loss-based probes and design more robust metrics for investigating the encoded knowledge in the existing language models.

\bibliography{anthology,custom}
\bibliographystyle{acl_natbib}
\appendix
\section{Structural Probe Analysis}
\label{sec:struct-probe}
\begin{figure*}[!ht]
\includegraphics[width=13cm]{./images/lagend-Prob-linguistic}
\includegraphics[width=13cm]{./images/lagend}
    \centering{
    \subfigure[Layer-wise analysis]{\includegraphics[width=7cm,height=5cm]{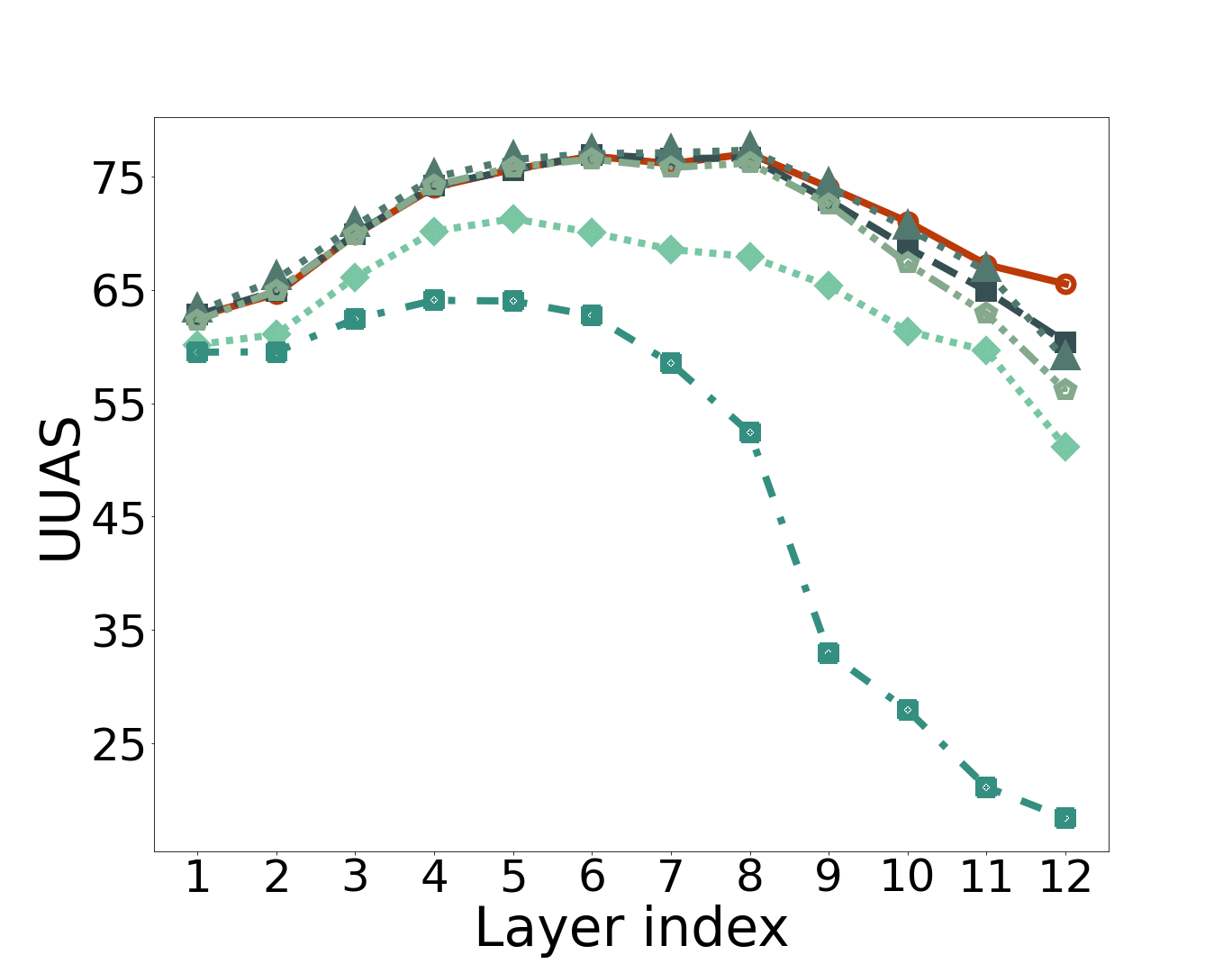} \label{fig:ling-struct}} 
    \subfigure[Data size impact]{\includegraphics[width=7cm,height=5cm]{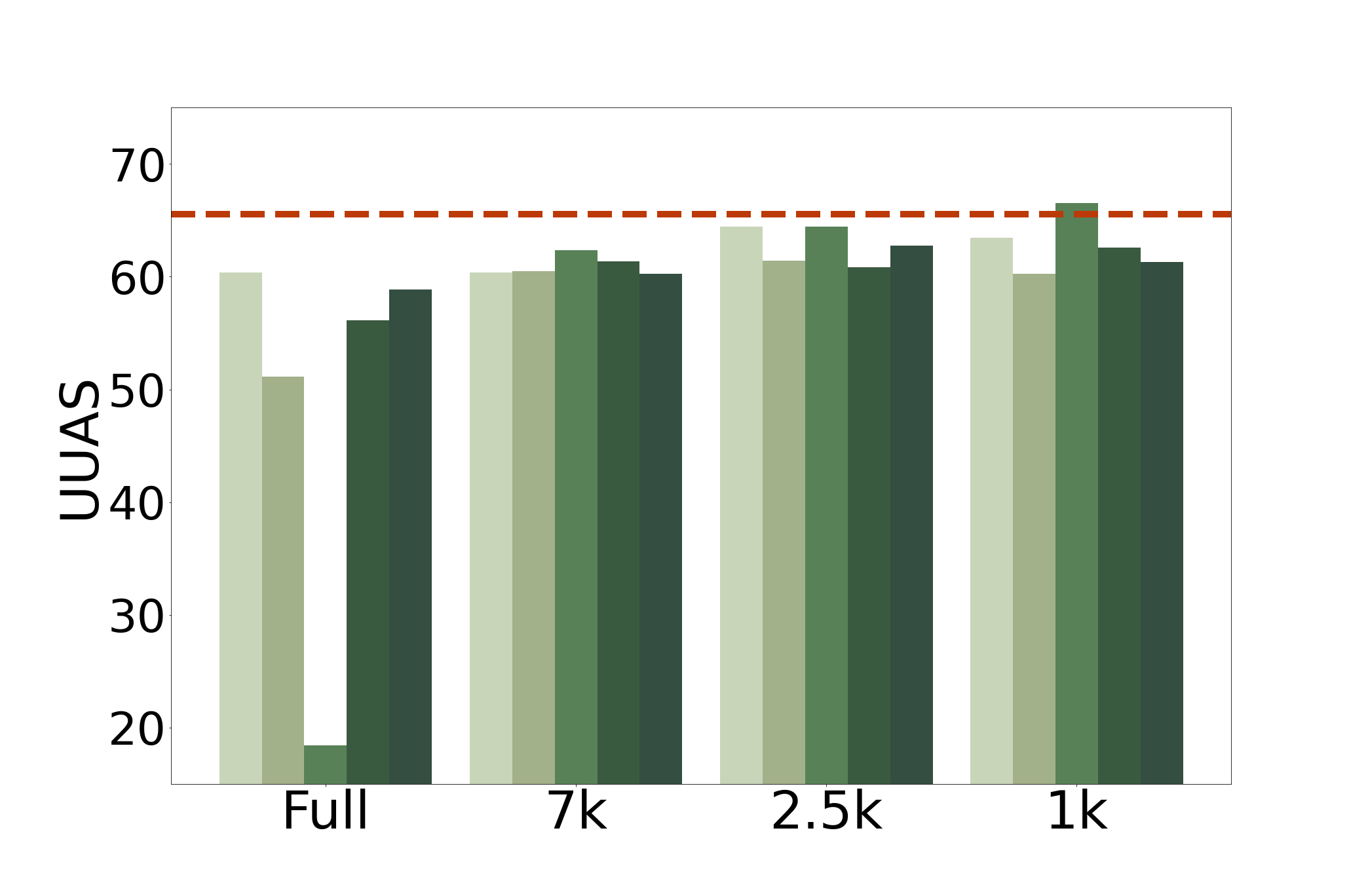}\label{fig:struct}}}
    \caption{(a) UUAS score of the structural probe on all layers of fine-tuned models. (b) The visualization of models' performance fine-tuned on the fixed-size training sets on the structural probe. The pre-trained BERT's performance is shown by the dashed red line.}
    \label{fig:structural-results}
\end{figure*}

We have also repeated our data size analysis experiment on the structural probe to show that our findings stand for different probes. Figure~\ref{fig:structural-results} confirms our conclusions drawn from Section~\ref{sec:datasize-analysis}, which denotes that data size affects the probing performance of fine-tuned models.

\section{Linguistic Knowledge Recoverability} 
\label{sec:appendixB}

\begin{figure*}
    \centering
    \includegraphics[width = 3.89cm]{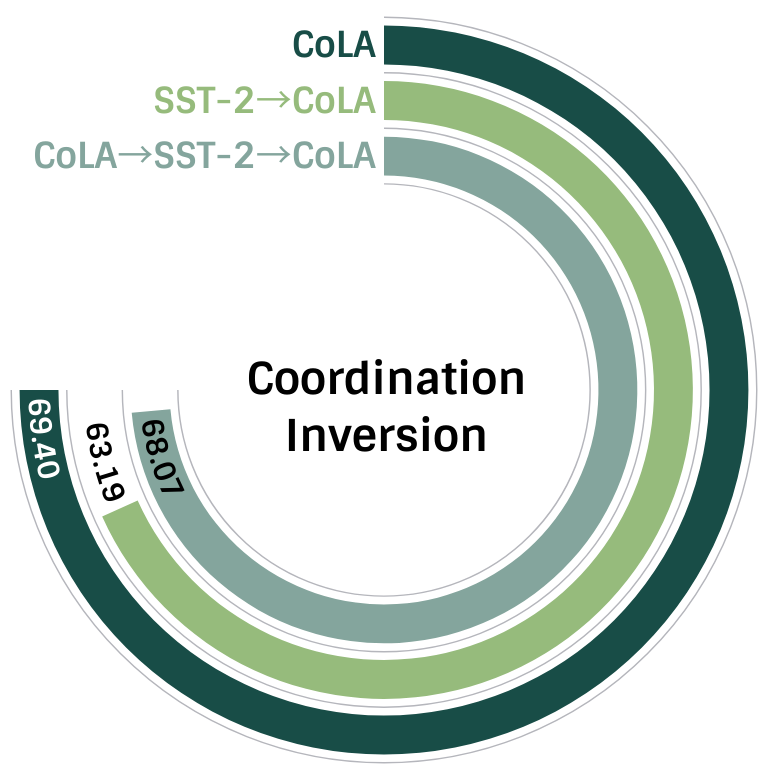}
    \includegraphics[width = 3.89cm]{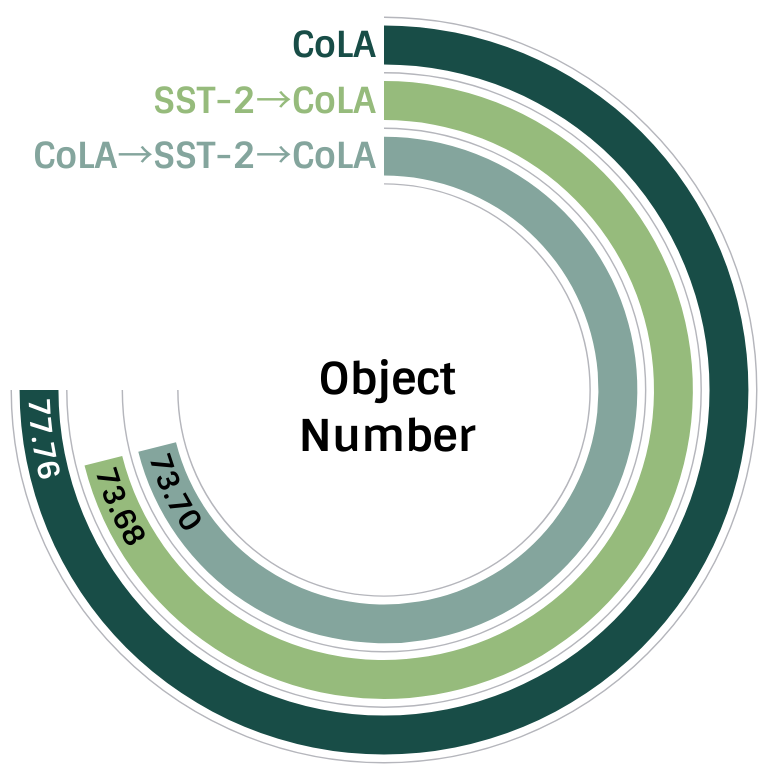}
    \includegraphics[width = 3.89cm]{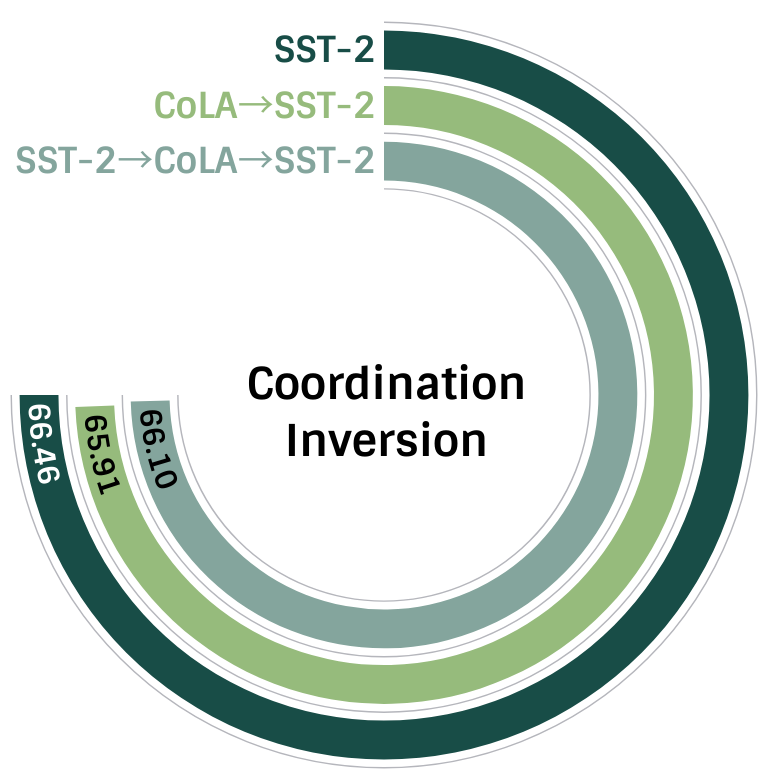}
    \includegraphics[width = 3.89cm]{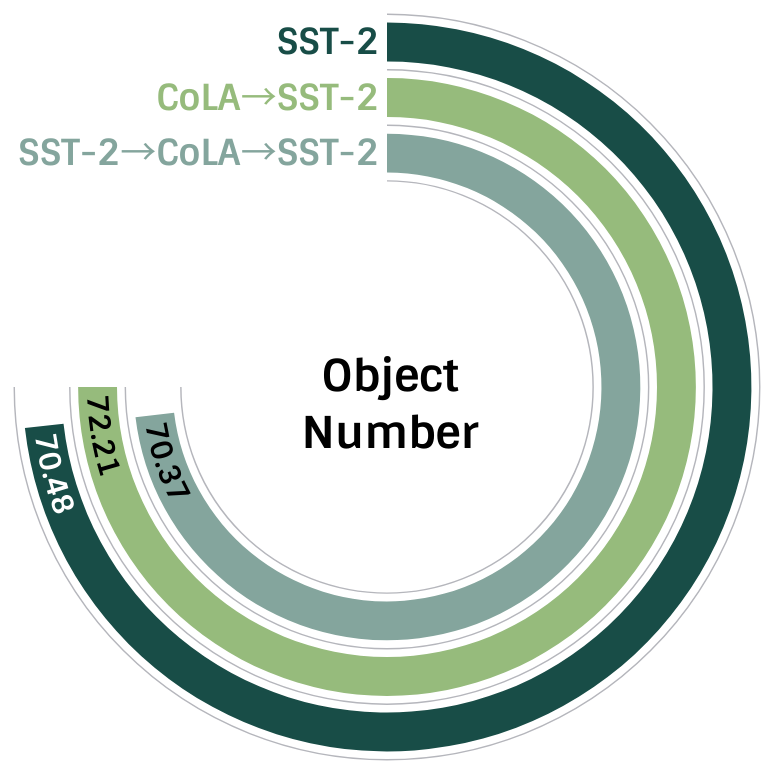}
    \includegraphics[width = 3.9cm]{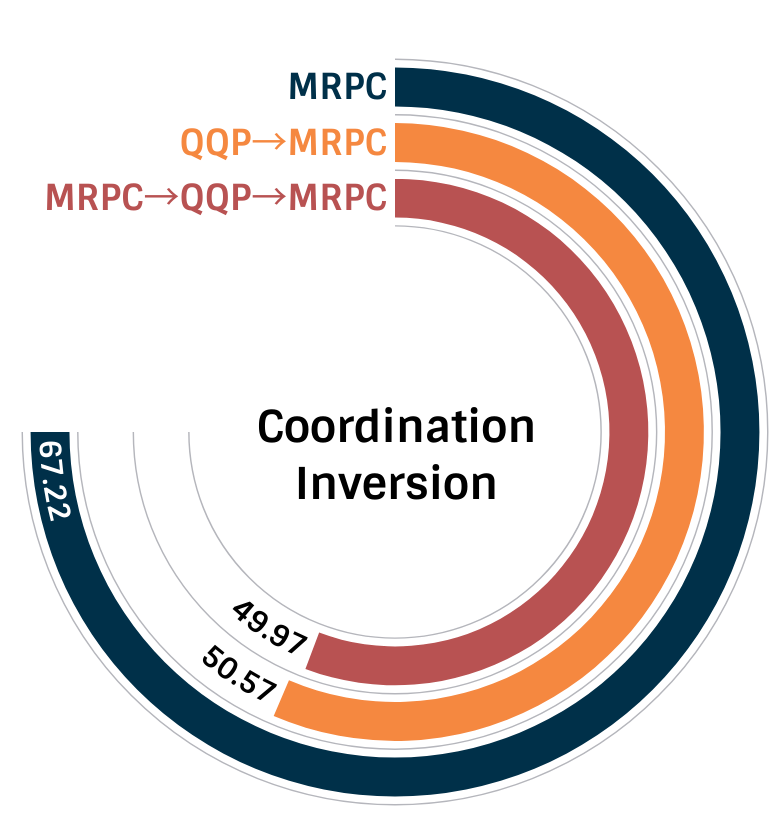}
    \includegraphics[width = 3.9cm]{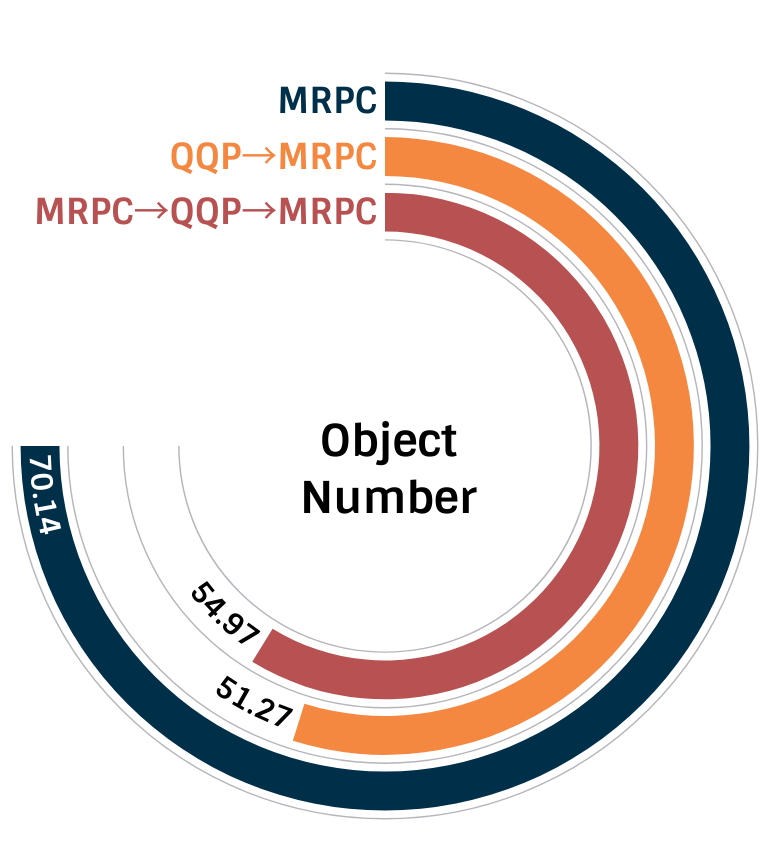}
    \includegraphics[width = 3.9cm]{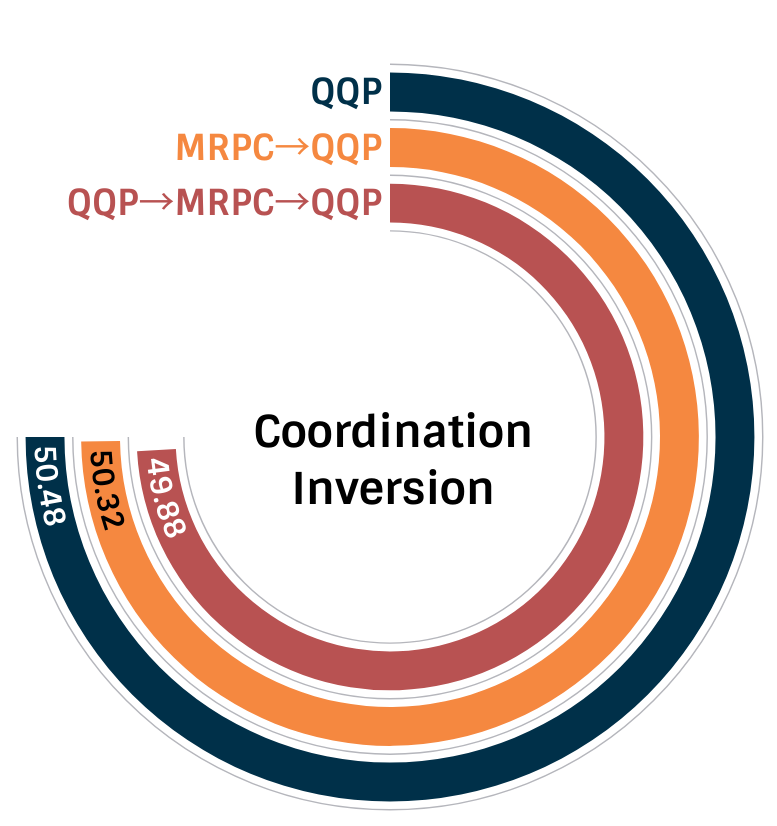}
    \includegraphics[width = 3.9cm]{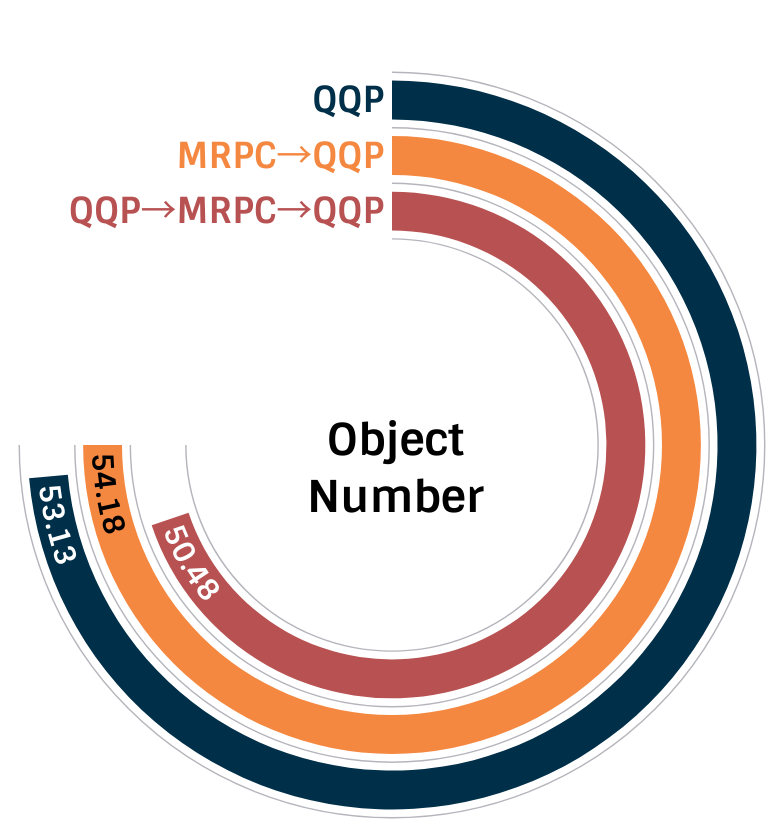}
    \caption{The performance of the models after being sequentially fine-tuned on different tasks. Three-quarters of a circle represents the maximum value and the outer circle is the baseline.}
    \label{fig:sequence-analysis-app}
\end{figure*}

\end{document}